 \documentclass[pmlr,twocolumn,10pt]{jmlr} 

\usepackage{booktabs}
\usepackage{hyperref}
\usepackage{float}
\usepackage{multirow}
\usepackage{makecell}
\usepackage[load-configurations=version-1]{siunitx} 

\theorembodyfont{\upshape}
\theoremheaderfont{\scshape}
\theorempostheader{:}
\theoremsep{\newline}

\newcommand{\kaiser}{Kaiser Permanente}

\jmlrvolume{LEAVE UNSET}
\jmlryear{2021}
\jmlrsubmitted{LEAVE UNSET}
\jmlrpublished{LEAVE UNSET}
\jmlrworkshop{Machine Learning for Health (ML4H) 2021} 

\title[SmartTriage]{SmartTriage: A system for personalized patient data capture, documentation generation, and decision support}

\author{%
 \Name{Ilya Valmianski} \Email{ivalmian@gmail.com}\\
  \Name{Nave Frost} \Email{Nave.X.Frost@kp.org}\\
  \Name{Navdeep Sood} \Email{Navdeep.X.Sood@kp.org}\\
  \addr Medical Informatics, Kaiser Permanente
  \AND
  \Name{Yang Wang} \Email{Yang.X1.Wang@kp.org}\\
  \addr Enterprise Architecture, Kaiser Permanente Digital
  \AND
  \Name{Baodong Liu} \Email{Baodong.X.Liu@kp.org}\\
  \Name{James J. Zhu} \Email{James.J.Zhu@kp.org}\\
  \Name{Sunil Karumuri}  \Email{Sunil.Karumuri@kp.org}\\
  \Name{Ian M. Finn}  \Email{Ian.M.Finn@kp.org}\\
  \Name{Daniel S.  Zisook} \Email{Daniel.S.Zisook@kp.org}\\
  \addr Medical Informatics, Kaiser Permanente
  }
  
\begin{document}

\maketitle

\begin{abstract}
Symptom checkers have emerged as an important tool for collecting symptoms and diagnosing patients, minimizing the involvement of clinical personnel. We developed a machine-learning-backed system, SmartTriage, which goes beyond conventional symptom checking through a tight bi-directional integration with the electronic medical record (EMR). Conditioned on EMR-derived patient history, our system identifies the patient's chief complaint from a free-text entry and then asks a series of discrete questions to obtain relevant symptomatology. The patient-specific data are used to predict detailed ICD-10-CM codes as well as medication, laboratory, and imaging orders. Patient responses and clinical decision support (CDS) predictions are then inserted back into the EMR. To train the machine learning components of SmartTriage, we employed novel data sets of over 25 million primary care encounters and 1 million patient free-text reason-for-visit entries. These data sets were used to construct: (1) a long short-term memory (LSTM) based patient history representation, (2) a fine-tuned transformer model for chief complaint extraction, (3) a random forest model for question sequencing, and (4) a feed-forward network for CDS predictions. In total, our system supports  337 patient chief complaints, which together make up $>90\%$ of all primary care encounters at \kaiser.
\end{abstract}
\begin{keywords}
Symptom checker, natural language processing, clinical decision support, Electronic Medical Record, health care, deep neural networks, transformers, decision trees, random forests
\end{keywords}

\section{Introduction}
\label{sec:intro}

In recent years, there has been a broad proliferation of online symptom checkers and medical triage solutions as many patients seek web-based medical guidance prior to engaging with their health care provider \citep{White2009, Semigran2015, Chambers2019, Dunn2020}. A typical symptom checker functions by asking patients a sequence of questions and generating a set of possible conditions and treatment options. Symptom checkers vary by their operational range -- some target specific populations (e.g. pediatrics) or disease states (e.g. influenza), while others have broader scopes (e.g. low acuity primary care) \citep{Kellermann2010, Price2013, BuoyHealth, Ada, Infermedica, BabylonHealth, MedialEarlySign}.  

The need for high quality symptom checkers grew with the COVID-19 pandemic, which caused a significant decrease in utilization of primary and urgent care \citep{Gelburd2020}. Suddenly, many millions of people are seeking diagnostic assistance online and up to 90\% of primary care visits have shifted to virtual care. The transition to (in some cases) majority virtual care represents an opportunity for greater symptom checker usage and an imperative for tight integration into the health care operational workflow.  

Symptom checkers with actionable outcomes can provide a host of benefits. Computer assisted triage of low acuity patients can decrease clinician patient load, while decision support for ordering of appropriate laboratory testing (including COVID-19 testing, \citep{Greenhalghm1182, Judson2020}), medications, and imaging studies, can eliminate unnecessary visits. With advanced decision support and documentation assistance, patient visits themselves are enhanced as the clerical load (e.g. documentation, coding, ordering) is decreased. 

Obtaining maximal benefit from symptom checkers requires a tight integration with the Electronic Medical Record (EMR). An integrated system can utilize knowledge gleaned directly from the patient and can also leverage historically relevant patient medical information. In addition, symptom checker output that is manipulable and represented in appropriate locations within the EMR can help facilitate clinical workflow and improve quality of care.

In this paper we present SmartTriage, a symptom checker that is tightly integrated with the EMR and which has a broad scope covering 337 presenting chief complaints. We use machine learning to bootstrap question sequencing and decision support predictions through training on two novel data sets. The first data set contains nearly one million patient generated free-text reasons-for-visit entries, each of which is mapped to a discrete chief complaint. The second data set is a collection of 25 million patient encounters including progress notes as well as discrete diagnoses and orders. 

Three machine learning models enable SmartTriage. Our chief complaint extraction model (CC model) uses a fine-tuned bi-directional transformer neural network (BERT) to process patient free-text reason-for-visit entries \citep{Devlin2019BERTPO}. A collection of over 3000 random forest models were trained to produce highly informative question sequences (QS model). Finally, a feed-forward network was created to produce clinical assessments (Assessment model), taking into account all historical data and all questions answered. The CC and the Assessment models use a long short-term memory (LSTM) based patient history representation \citep{Hochreiter1997}. In this paper we also emphasize the role of medical history on the model performance, as it is valuable but often unutilized information source about the patient.

SmartTriage has been piloted in a single urgent care facility and is currently being piloted across the \kaiser Southern California Region in virtual visits (phone, video) as a pre-visit symptom checker. Details about the production integration can be found in Appendix \ref{apd:prod}.
\vspace{-1mm}
\section{Methods}
\vspace{-1mm}
\subsection{Content authoring}
\label{sec:content}
Clinical concepts from the UMLS are by their nature very detailed and clinically focused \citep{Bodenreider2004}. The level of detail often implies that multiple concepts exist to capture the same idea in different contexts. \figureref{fig:concept_lifecycle} shows three concepts derived from a clinical narrative including the phrase ``c/o pain after eating." The concepts describe gastric contents backing up into the esophagus - ``Heartburn” as the sign or symptom, ``Gastroesophageal reflux” as the finding on exam, and ``reflux” as a pathological process in general. Although there are context clues in medical notes, clinicians will often use these phrases or variants to imply the same idea, and our NLP engine will produce one or more of these concepts for the same span of clinical text.

\begin{figure}[!htb]
   \floatconts{fig:concept_lifecycle}
   {  \caption{Diagram of concept life cycle. Extracted raw UMLS concepts are further processed and grouped into a single ST-CMS concept. For each ST-CMS concept we author separate physician and patient-friendly descriptions.}}
   {
  \includegraphics[width=0.7\linewidth]{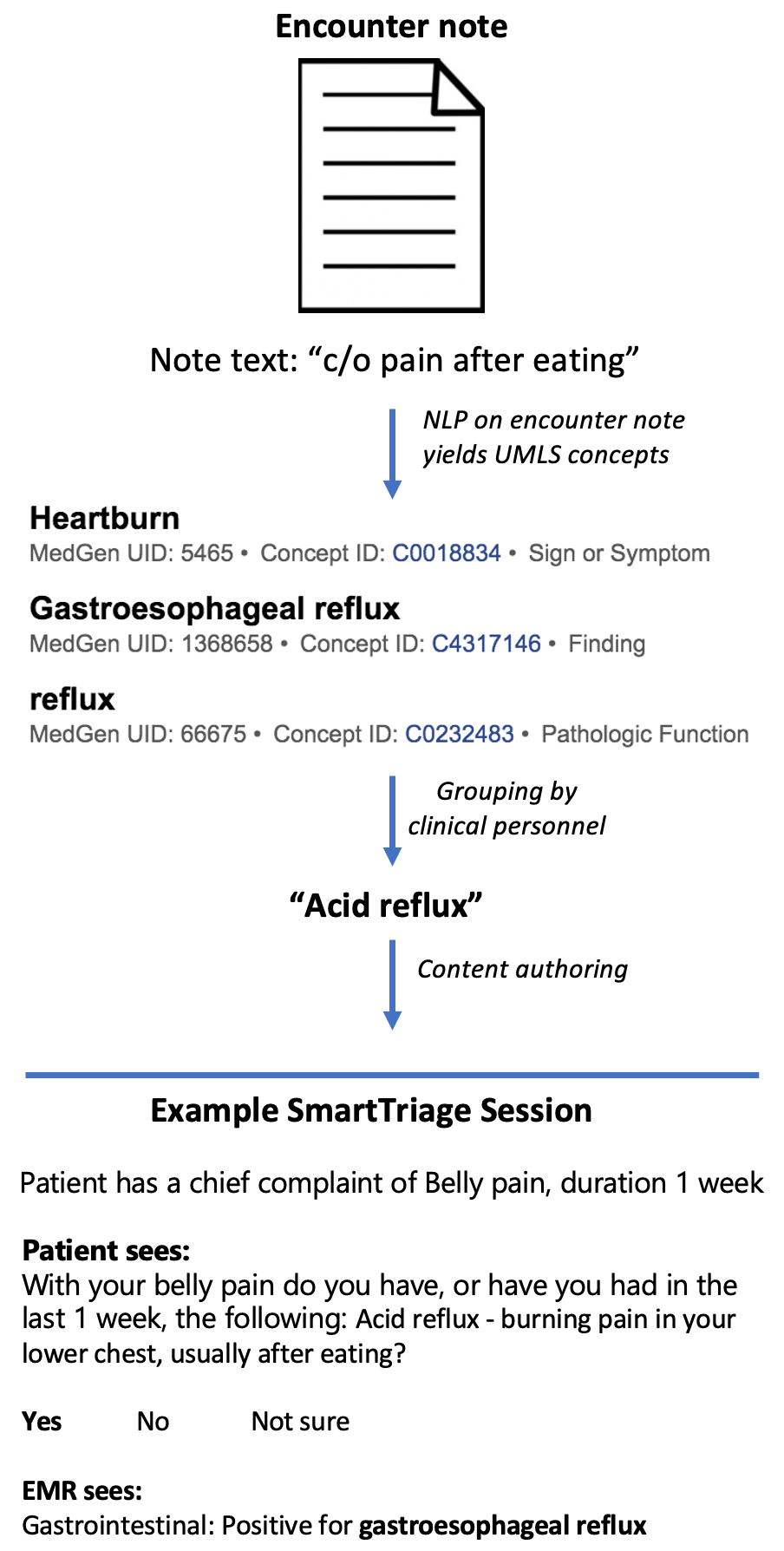}

 }
\end{figure}

In order to prepare a high quality training data set for our QS and Assessment models, we employed many man-hours of clinical review to collapse all of the items found by our NLP engine into meaningful ``main idea" groups. In large part, the groups contained perfectly synonymous concepts; in some cases, however, we grouped items that were not perfectly synonymous but implied the same concept. \figureref{fig:concept_lifecycle} shows the three UMLS concepts ``Heartburn", ``Gastroesophageal reflux" and ``reflux" grouped together into a container ``Acid reflux" concept. Our pilot system contains more than $3\textnormal{K}$ distinct SmartTriage Content Management System (ST-CMS) concepts. 

SmartTriage is both patient and provider facing, and thus we had the dual challenge of authoring our grouped concepts such that they would be understandable to patients as questions and to providers as clinical concepts. An example of authored patient and provider facing content can be seen in the ``Example SmartTriage Session" portion of \figureref{fig:concept_lifecycle}. Here we combine a canned phrase with patient friendly language to inquire about the symptom, and serve back to the EMR an appropriate clinical phrase and sense (positive or negative) to capture the patient's response. 

Showing questions to patients and bringing patient answers into the EMR to be placed into a progress note requires tagging each question/concept with metadata describing how it should be displayed to patients and where it should go in the clinical note. \figureref{fig:cmsview} shows the system we built to manage all of our content called the Content Management System (CMS), including concept groupings, patient facing text, provider facing text, and question metadata. Groupings are managed by establishing a parent-child ``synonym" based relationship with functionality to add, remove, and change the parent. Each concept may have a range of response types, from simple yes/no to multi-select to free text, and the CMS allows that to be configured. Authoring questions may include pre-text (a stem), the question itself, and post-text, all of which are editable. Metadata or question ``attributes" control where in the clinician documentation the patient responses will go and how the patient's response is meant to be interpreted by the ML components of the SmartTriage system, a parameter called ``response evaluation type." Finally, the CMS creates a preview of the question as it would be seen by the patient for validation of phrasing and question configuration.

\subsection{Chief Complaint and Assessment models}
\label{sec:nn_method}

The neural networks used in the Chief Complain and Assessment models were implemented using Tensorflow 2.3 \citep{tensorflow}. For all models we used Adam optimizer \citep{Kingma2015}. Model architecture are schematically represented in \figureref{fig:neural_net_components}. Medical history component uses the same architecture for both assessment and chief complain models but are trained separately with each model.

\paragraph*{Medical history input cleaning} Medical history input data was limited to items occurring at least 5000 times in the Encounters data set.

\paragraph*{Chief Complain model target cleaning} The discrete chief complaints selected by clinical personnel were further manually reviewed and grouped into 337 discrete chief complaints which were used as targets for the chief complaint model as well as inputs for the Question Sequencing and Assessment models.

\paragraph*{Assessment model target cleaning} For each target value $t$ and chief-complaint $c$, the Bayesian lift of $t$ appearing given $c$ was calculated, i.e. $\textnormal{lift}[t | c] = \frac{\mathbb{P}[t | c]}{\mathbb{P}[t]}$. Targets with a value $\textnormal{lift}[t | c] \leq 2$ were filtered out.

\paragraph*{Assessment model loss weighting} Loss for each output layer was weighted differently by performing a random search, limited to 10 combinations, over a $5\%$ sample of the data, where each of the loss terms ($\mathcal{L}_{\text{diag}}$, $\mathcal{L}_{\text{med}}$, $\mathcal{L}_{\text{lab}}$, and $\mathcal{L}_{\text{img}}$) was weighted by $w \in \{0.5, 0.9, 1.0, 2.0, 3.0\}$. The different weight combinations were sorted by the PR-AUC of all four output types, and the combination with highest rank in its worst performing output type was selected. The highest performing weighting function found was $\mathcal{L} = 3 \cdot \mathcal{L}_{\text{diag}} + \mathcal{L}_{\text{med}} + 0.9 \cdot \mathcal{L}_{\text{lab}} + 0.9 \cdot \mathcal{L}_{\text{img}}.$
\vspace{-4mm}
\subsection{Question Sequencing model}
\vspace{-1mm}
\paragraph*{Data cleaning} Similarly to the assessment model, question selection uses Bayesian lift to filter the target prediction. For ICD-10-CM diagnoses we calculated the Bayesian lift that a diagnosis $t$ given the chief-complaint $c$i.e. $\textnormal{lift}[t | c, a, s] = \frac{\mathbb{P}[t | c]}{\mathbb{P}[t]}$. Target with $\textnormal{lift}[t | c, a, s] \leq 2$ were filtered out. Furthermore, for every cohort of chief complaint, age-bin, and sex, we filtered out diagnoses and concepts that appeared fewer than 100 times. For each cohort, we further limited the input to the 1000 most-common concepts, and limited the prediction targets to the 50 most-common diagnoses (note that the assessment model, which is used for the final differential diagnosis prediction, does not have this limit). 
\vspace{-1mm}
\paragraph*{Training ML question sequencing} We use the encounter data set (Section \ref{sec:enc_demo}) to train the ML question sequencing model. For maximal interpretability, appropriateness, and ability to handle large feature spaces ($>1000$ features per model) we chose to train a random forest model for each combination of chief complaint, age-bin, and biological sex. The models were trained independently using ST-CMS concepts and assertions (derived from NLP on progress notes) as features and relevant ICD-10-CM diagnoses as labels.  ICD-10-CM diagnoses was chosen as a prediction target because they offer a broad proxy for patient disease states and are coded by medical professionals for every encounter in our data set. However, ICD-10-CM also introduce some bias as coding behaviors may be influenced by reimbursement dynamics and compliance to certain standard of care guidelines (e.g. a patient with cough and other upper respiratory issues may be coded as R05 or J06.9 depending on whether the treating physicians wants to prescribe antibiotics). The random forest models are trained with 100 estimators, feature sub-sampling of 10\%, and a max depth of 15. The model was implemented using sci-kit learn 0.23 \citep{scikit-learn} RandomForestClassifier algorithm.

\vspace{-1mm}
\paragraph*{Selecting next ML question} At inference time we take a vote from the random forest using \algorithmref{algo:qs_ml}. For every tree in the forest we traverse the nodes until we find a ST-CMS concept for which there is no assertion and extract the relevant question $q_i$. For each question we calculate a weight, $w_{q_i}$ that is the number of trees that are split on a concept asked by $q_i$. The candidate question is validated using two rules. First, we check in the knowledge base (KB) if there is another question that should be asked  (for example, for chief complaint ``foot problem" before asking ``How long have you had ankle pain?" we check for the presence of ``ankle pain"). If the answer to the prerequisite question is asserted as ``absent" then the weight of the target question is set to 0. If the answer to the prerequisite question is asserted as ``certain" then the question will not be modified. If the prerequisite question has not been answered then an additional boolean flag decides if the prerequisite question is first asked (e.g. question ``How long have you had ankle pain?" is replaced by ``Do you have ankle pain?") or if the question can be asked without changes. We then sum the weights of all questions that are mapped to the same prerequisite. Second, we check candidate questions for ``inappropriate" questions (e.g. some insomnia sequences that include sleep apnea may result in questions related to coughing that physician subject matter experts decided should not be asked). The highest weighted question at the conclusion of all ML and rule-based steps is then asked.

\vspace{-2mm}
\section{Results}
\label{sec:results}

\vspace{-1mm}
\subsection{Data sets}
\label{sec:datasets}

SmartTriage is built on historical EMR data from two novel data sets.  The first is based on a collection of patient reasons-for-visit entries and the second is based on a collection of primary care encounters. \figureref{fig:dataset_schematic} shows how the data sets were acquired. 

When a patient schedules an appointment online they are prompted to enter a free-text reason-for-visit. At the time of the encounter, a discrete chief complaint is chosen by the intake personnel. The combination of patient entered text, the discrete chief complaint(s), and 1 year of patient history from the EMR form the Patient-Reason for-Visit data set, which contains 907,170 encounters involving 595,692 unique patients  (see Appendix \ref{sec:cc_demo}).  

During a primary care encounter a progress note is typically generated containing the patient's history of present illness, past medical history, a review of systems, a physical exam, and an assessment and plan. The encounter may also result in a set of discretely coded outcomes such as ICD-10-CM codes for diagnoses, medication prescriptions, and laboratory or imaging orders. We use an in-house developed natural language processing (NLP) engine \citep{Torii2015, Fan2013} to extract Unified Medical Language System (UMLS) clinical concepts from progress notes \citep{Bodenreider2004}. The UMLS concepts were then further processed using our content authoring process (see \ref{sec:content}) and mapped to patient and physician facing text. The extracted discrete concepts, encounter outcomes, and 1 year of patient history from the EMR form the Encounters data set, which contains 25,047,535 encounters involving 4,110,052 unique patients (see Appendix \ref{sec:enc_demo}).

\vspace{-1mm}
\subsection{Machine learning components}
\label{sec:ml_comp}

\begin{figure}[]
\centering
\begin{minipage}{0.8\linewidth}
\begin{algorithm}[H]

	\caption{Random forest voting}
	\label{algo:qs_ml}
	\setcounter{AlgoLine}{0}
	\LinesNumbered 
	\setcounter{AlgoLine}{0}
	\BlankLine
	\LinesNumbered 
	\SetKwInOut{Input}{Input}
    \SetKwInOut{Output}{Output}
	
	\Input{Feature set $X$, random forest $F$}
	\Output{Question candidates $Q$ and their weights $W$}
	\BlankLine

	Initialize $Q,W \leftarrow \emptyset$ \;

	\ForEach{\textnormal{Decision tree} $e \in F$}{
	
		Traverse $e$ using features $x \in X$ until reaching node $n$ that is either a leaf node or cannot be split using $X$\;
		
		\uIf{$n$ \textnormal{is not a leaf node}}
			{		 
				 Let $q$ be question to obtain feature $x$ that splits $n$\;
				 
				 $Q \leftarrow Q \cup \{q\}$
			}

	}
	
	Initialize $\tilde{Q} \leftarrow \textnormal{unique}(Q)$\;

	\ForEach{$\tilde{q_i} \in \tilde{Q} $}{
		$w_{\tilde{q_i}} \leftarrow \underset{j}{\sum} \delta(q_j, \tilde{q_i}) $\;
		
		$W \leftarrow W \cup \{w_{\tilde{q_i}}\}$
	}
	
	\Return $\tilde{Q},W$\;
	
\end{algorithm}
\end{minipage}
\end{figure}

There are three machine-learning-based components in SmartTriage. In order of operation, the Chief Complaint (CC) model takes patient history as well as a patient entered free-text reason-for-visit and generates predictions for the discrete chief complaint. The Question Sequencing (QS) model uses the discrete chief complaint, patient history, and patient responses, to interactively ask a series of questions with discrete response choices. Finally, the Assessment model employs all of the collected information to generate predictions for diagnoses as well as medication, laboratory, and imaging orders. 
\vspace{-1mm}
\subsection{Chief complaint extraction}
\label{sec:cc_model}

\paragraph*{Architecture} A feed-forward neural network was constructed to combine pre-trained free-text embeddings with patient age (discretized into age groups, see groups in Table \ref{table:datasetAgeDistribution}), biological sex, and patient medical history. As multiple discrete chief complaints can be extracted from a given patient-generated reason-for-visit, the output of the model is a one-vs-rest classification for 337 chief complaints. We compared several different text embedding approaches and layer size parameters, for detailed results see Table \ref{table:CCEmbedPerformance}. The best performing embedding was a custom fine-tuned BERT model,  Chief Complaint (CC) BERT, which was trained on a portion of the CC data set (anonymous citation). For detailed network architecture see \figureref{fig:neural_net_components}.

\begin{figure*}[!htb]
     \centering
    \includegraphics[width=0.8\textwidth]{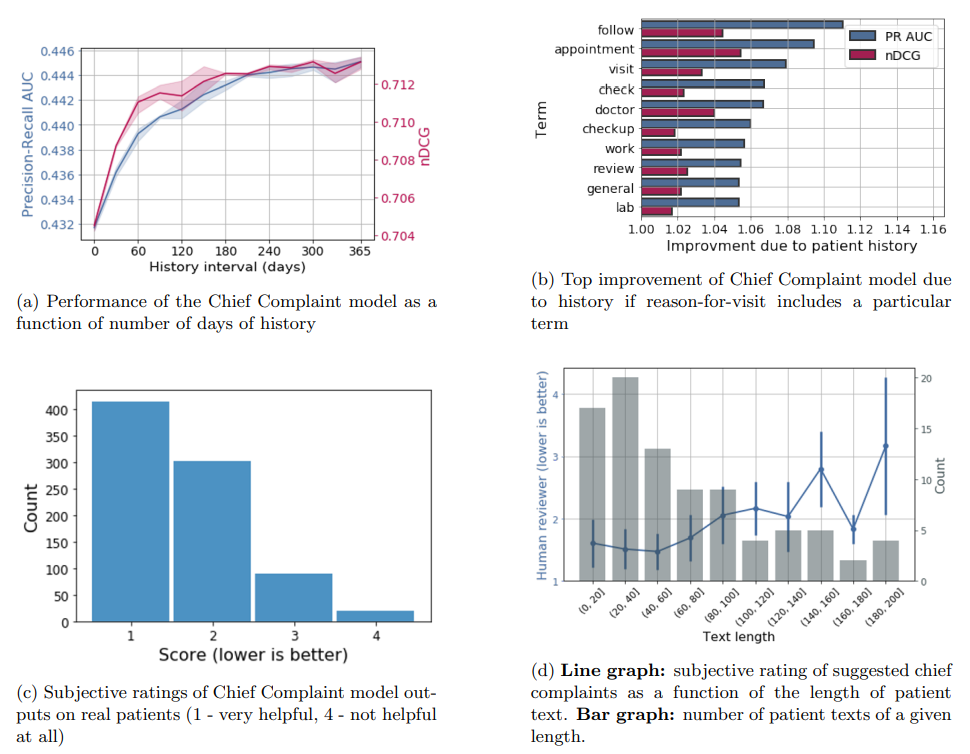}
      
         \caption{Chief Complaint model performance}
        \label{fig:cc_perf_w_len}
\end{figure*}

The CC model utilizes patients medical history from encounters over the preceding 365 days. The history is composed of 4 different channels: previous diagnoses (1,497 distinct values), prescriptions (1,278), procedures (758), and chief complaints (954). Data from each historic encounter were placed in the four channels and channels with no relevant information were filtered out. Each channel contained the last 8 encounters with relevant information sorted in ascending order by the encounter date. We zero padded channels in cases where there were data from fewer than eight encounters. We used a single layer LSTM to combine medical history into a fixed sized representation (see \figureref{fig:cc_perf_w_len}a) \cite{pham2017predicting, rajkomar2018scalable}.  The outputs of the recurrent layers and the demographic embeddings were then combined with the pre-trained free-text embeddings.

\begin{table*}[h!]
	\caption{Chief complaint model performance using various text embeddings}
	\centering
	\begin{tabular}{llll}
		\midrule
		& \textbf{PR AUC}        & \textbf{ROC AUC}       & \textbf{nDCG} \\
		\toprule
		TF-IDF baseline & 0.4365±0.0002 & 0.9203±0.0008 & 0.7070±0.0002    \\
		BERT base       & 0.4210±0.0004 & 0.9248±0.0007 & 0.6970±0.0003    \\
		BioBERT         & 0.4194±0.0017 & 0.9245±0.0005 & 0.6962±0.0009     \\
		Clinical BERT   & 0.4254±0.0011 & 0.9248±0.0003 & 0.6998±0.0010     \\
		MD BERT         & 0.4411±0.0006 & \textbf{0.9260±0.0003} & 0.7102±0.0003     \\
		CC BERT & \textbf{0.4451±0.0005} & 0.9254±0.0004 & \textbf{0.7132±0.0004}     \\
		\bottomrule   
		\label{table:CCEmbedPerformance}
	\end{tabular}
\end{table*}

\paragraph{Text embedding}
We  compared  6 different text embedding models (see Table \ref{table:CCEmbedPerformance}).
Pre-trained models for text embeddings were obtained from three publicly available sources: BERT-base \citep{devlin_bert:_2019}, BioBERT \citep{lee_biobert:_2019}, and Clinical BERT \citep{alsentzer_publicly_2019}. In addition, we fine-tuned a MD BERT model starting with Clinical BERT, using both the masked language model and the next sentence prediction task with 100 million sentence pairs extracted from patient progress notes (from encounters unrelated to the Chief Complaint Dataset). We also fine-tuned Chief Complaint (CC) BERT on the training portion of the CC Dataset, starting from MD BERT, using the masked language model prediction task. To generate BERT-based embeddings, we mean-pooled the last 4 layers, yielding an embedding size of 3,072. 

As a baseline, we also trained a version of the CC model using TF-IDF embeddings. Pre-processing before TF-IDF vectorization included: (1) transforming the text to lowercase; (2) removing standard English language stop-words; (3) replacing all numbers with the generic ``\#” symbol. The Penn Treebank (PTB3) tokenizer  implemented in NLTK \citep{bird_natural_2009} was used to tokenize the text, and unigrams and bigrams were then computed. TF-IDF vectors were created consisting of values for the 50,000 most common n-grams, with at most 0.5 document frequency in the training set.

For each model, we generated 5 estimates by training in 5-fold cross validation. Resulting average and standard deviation micro Precision Recall Area Under the Curve (PR AUC), micro Receiver Operating Characteristic Area Under the Curve (ROC AUC), and Normalized Discount Cumulative Gain (nDCG) values can be found in Table~\ref{table:CCEmbedPerformance} \citep{Zhang2009, nDCGmetric}. In terms of PR AUC and nDCG the best performing embedding was CC BERT, with statistical significance ($p<0.001$). MD BERT performed slightly better than CC BERT in terms of ROC AUC, though the difference between the methods was not statistically significant.

Finally, for the model that used CC BERT embeddings we performed hyper-parameter tuning to identify proper dropout rate and size for the feed-forward layer after contactenating patient text, demographic embedding, and medical history embedding, see Table \ref{table:CCHyperPerformance}. We found that while increasing the layer size somewhat improved performance, the improvements were minimal for layers larger than 500 units, thus we used the 500 unit layer with dropout of 0.5 in our final product.

\paragraph{Medical history contribution} We examined the effect of including patient medical history on CC model performance by training models with up to 1 year of patient history. We found that both PR AUC and nDCG increase slightly with inclusion history (PR AUC without history 0.432, with one year history 0.445, nDCG without history 0.705, with history 0.713; see \figureref{fig:cc_perf_w_len}) by training a baseline model without the medical history input and embedding layers. The impact of history was skewed, however, which was revealed by looking at PR AUC changes achieved for complaints containing free-text terms with prevalence greater than $1\%$ (73 terms total). For each of these terms we bucketed chief complaints containing the relevant term and compared the PR AUC and nDCG of both models. For 71 of the groups the model with history over-performed in terms of PR AUC, and 66 of the groups over-performed in terms of nDCG. The highest impact of patient history was seen in free-text reason-for-visit containing words relating to previous encounters, such as ``follow'', ``appointment'', ``visit'', and ``check''  (see \figureref{fig:cc_perf_w_len}), where improvement is measured as $\frac{\text{Metric with history}}{\text{Metric without history}}$.

\paragraph{Subjective reviewer evaluation} Because our system produces chief complaints that are then presented back to patients, we needed to ensure that only appropriate chief complaints are predicted. To do this we performed a manual review on predictions for 275 patient generated reason-for-visit. We employed 3 reviewers which were presented with the patient text and SmartTriage generated chief complaint predictions. The reviewers were presented up to 5 predictions with a minimum model output score of 0.05.  They were prompted to holistically evaluate the helpfulness of the predictions with the following scoring rubric (Very helpful = 1, Helpful = 2, A little helpful = 3, Not helpful = 4). The average rating over all patient reason-for-visit was $1.7\pm0.8$ with root mean square inter-reviewer disagreement of 0.5. Of the 825 reviews (275 x 3) only 20 were marked as ``Not helpful" (see \figureref{fig:cc_perf_w_len} for full breakdown). We also observed model performance as a function of text length (\figureref{fig:cc_perf_w_len}), where we found that the score deteriorates for reasons $> 100$ character long. One explanation for the deterioration is that long patient reason-for-visit often contain multiple potentially unrelated complaints.

\vspace{-3mm}
\subsection{Question sequencing}
\label{sec:qs_model}

Question sequencing occurs after the chief complaint is established. By asking a series of structured and ML-derived questions, SmartTriage captures patient data that can be used to help generate the progress note. Patient responses are also fed as inputs to the assessment model. All questions and answers displayed to the patient are mapped to ST-CMS concepts (see Section \ref{sec:content}). Questions may be multi-select, single-select, yes/no, drop-down (for duration questions), and free-text. Each answer has an assertion associated with it that is typically drawn from \{``certain", ``absent", ``unsure"\}. For duration and severity questions the assertions are ordinal (number of days for duration and integer-scale between [0,10] for severity). 

Individual questions in SmartTriage are required to be interpretable  and appropriate. Interpretable means that we can explain why each question was asked based on previous answers and patient medical history. Appropriate means that SmartTriage should not ask questions whose answers could be inferred either from patient history or from the set of previous patient responses. To satisfy the appropriateness requirement we use a combination of rules and machine learning, both of which have available information on the chief complaint, previously answered questions, patient demographics, and 6-months worth of patient medical history (previous chief complaints, diagnoses, laboratory results, and medications; note that this is different from the CC and the Assessment models that use 1 year worth of patient medical history).

\vspace{-1mm}
\paragraph{Architecture} Schematic of the QS algorithm is presented in \figureref{fig:question_selection}. At every step the asserted ST-CMS concepts (features) are used to infer additional asserted concepts via a rules engine. For the full set of 337 covered chief complaints, the initial rule-driven questions are stereotyped to collect a thorough chief-complaint-relevant history of present illness (HPI) with relatively minimal hard-coded branching to cover biological sex differences, timing variations, problem locations, etc. For a subset of 269 chief complaints, after the rule-based HPI questions are exhausted ML-derived questions are asked. The ML questions are subject to additional ``fixer" rules, and can extend the question sequence up to a set number of total questions.

\begin{figure}[!htbp]
\floatconts
{fig:qse_hist}
{\caption{Question sequencing performance evaluated on ICD-10-CM prediction task.  }}
{
    
         \includegraphics[width=0.4\textwidth]{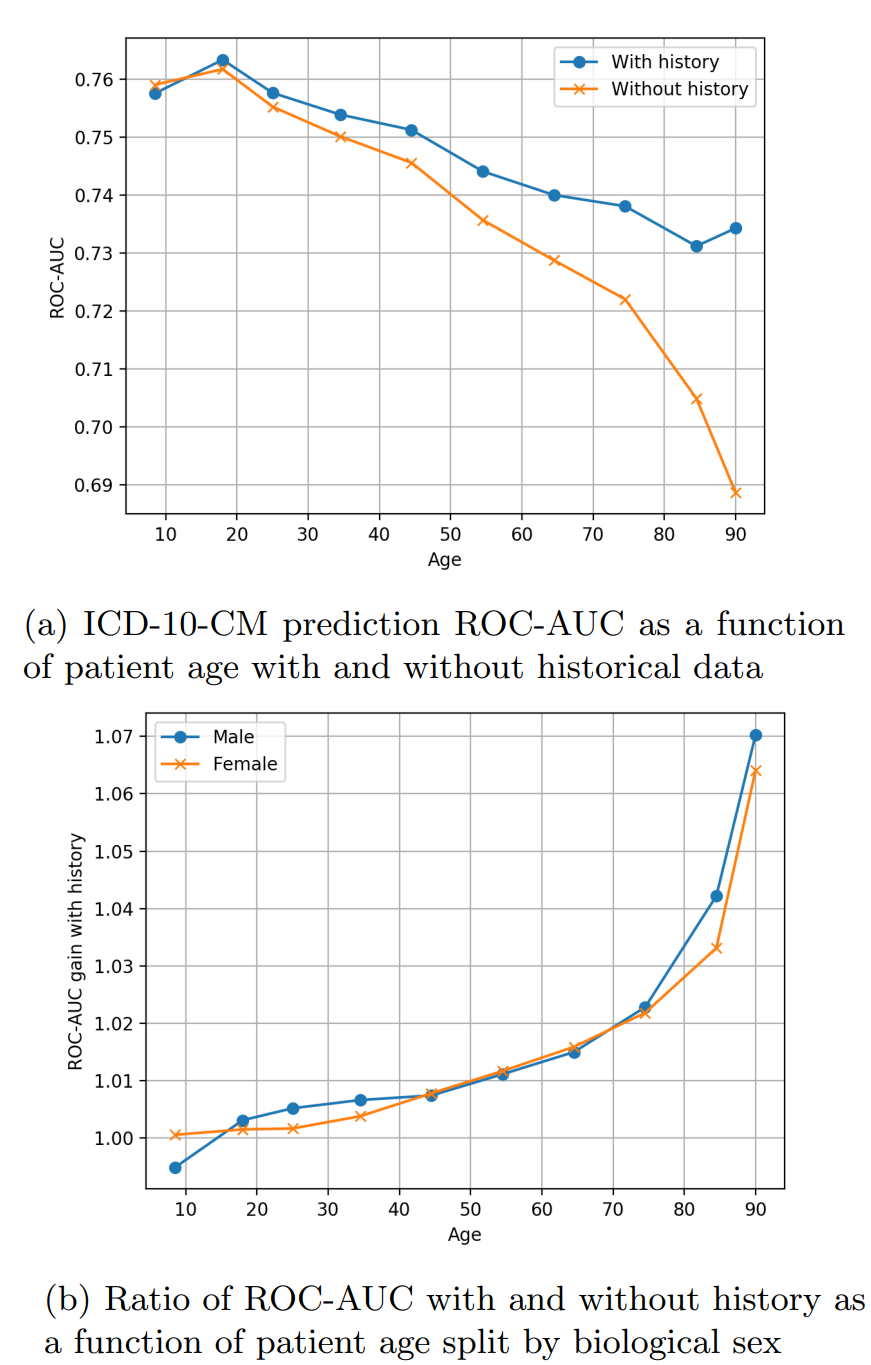}

}
\end{figure}
\vspace{-3mm}
 \paragraph{Medical history contribution} We trained the random forest models with and without patient medical history. Both models achieved a very similar micro ROC AUC with respect to ICD-10-CM predictions ($0.747$ with history vs. $0.737$ without history, $1.3 \% $ difference). However, we found that patient medical history disproportionately affects performance for older patients, with up to $7 \% $ higher performance for patients older than 90 years old, see \figureref{fig:qse_hist}. We also compared the effect of history as a function of biological sex, and found that both male and female patients have a similar dependence on age, see \figureref{fig:qse_hist}. To understand better what drives the benefit provided by history, we examined the ratio in ICD-10-CM prediction micro ROC AUC with and without history for different chief complaints. We  found that chronic conditions tend to have the highest improvement when including patient medical history information (e.g. ``blood sugar problem" and ``diabetes care management" improved by $\sim 20 \% $; ``hypothyroidism", ``carpal tunnel syndrome", and ``chronic pain" by $\sim 15 \% $). Among acute diseases the ``bronchitis" and the ``pneumonia" chief complaints had the highest benefit from patient history at $\sim 10 \% $. Overall, medical history was more impactful with older patients because they had more available patient medical history  and a larger chronic disease burden.
\vspace{-1mm}
\subsection{Patient assessment}
\label{sec:assessment_model}

The Assessment model is the final ML model in the SmartTriage workflow, and it uses all of the collected information to predict the encounter diagnoses (1,240 distinct values) as well as medications (122), laboratory (291), and imaging (208) orders. 
\vspace{-1mm}
\paragraph*{Architecture} Inputs to the Assessment model are comprised of the discrete chief complaint, patient demographic data, answers from the QS model, and patient medical history. The Assessment model itself is a feed-forward neural network (\figureref{fig:assessment_model_arch}) with an LSTM based patient history embedding (\figureref{fig:assessment_model_arch}). Each of the inputs is embedded separately, and then the embeddings are concatenated together. The combined embbeding is passed through seven ReLU dense layers of size 1024 with skip connections. The representation returned from the second layer is used by the four one-vs-rest logistic regression output layers. Comparisons of different model types and architectures are available in Table \ref{table:assessment_architecture}. 

After a patient session with SmartTriage, the patient's responses and history are provided to the Assessment model and diagnoses, medications, labs and imaging orders are inferred. These inferences are fed back into the EMR and appear as decision support suggestions in the clinician workflow we designed for SmartTriage. If a clinician agrees with a particular diagnosis suggested, they may toggle it and the diagnosis will automatically be added to the set of current encounter diagnoses. We do not currently support direct ordering of medications, labs, or imaging studies but that functionality is planned to be implemented in the near future.

\begin{table*}[]
\floatconts
{table:assessment_architecture}
{\caption{Comparison of assessment model PR AUC, ROC AUC, and nDCG with different model types}}
{
\begin{tabular}{lllll}
\toprule  
\makecell[l]{\textbf{Metric}} &
  \makecell[l]{\textbf{Target}} &  \makecell[l]{\textbf{Logistic} \textbf{regression}} &  \makecell[l]{\textbf{1 hidden} \textbf{layer}} &  \makecell[l]{\textbf{7 hidden} \textbf{layers}} \\ \midrule
\multirow{4}{*}{PR AUC}  & Diagnoses   & 0.258 & 0.428                           & \textbf{0.432} \\
                         & Medications & 0.165 & 0.317                           & \textbf{0.325} \\
                         & Labs        & 0.178 & 0.264                           & \textbf{0.265} \\
                         & Imaging     & 0.068 & 0.187                           & \textbf{0.190} \\ \midrule
\multirow{4}{*}{ROC AUC} & Diagnoses   & 0.894 & 0.977                           & \textbf{0.979} \\
                         & Medications & 0.946 & 0.988                           & \textbf{0.989} \\
                         & Labs        & 0.925 & 0.943                           & \textbf{0.944} \\
                         & Imaging     & 0.792 & 0.871                           & \textbf{0.875} \\ \midrule
\multirow{4}{*}{nDGC}    & Diagnoses   & 0.513 & 0.596                           & \textbf{0.597} \\
                         & Medications & 0.210 & 0.234                           & \textbf{0.235} \\
                         & Labs        & 0.254 & 0.274                           & \textbf{0.275} \\
                         & Imaging     & 0.110 & \textbf{0.124} & \textbf{0.124} \\ \bottomrule     
\end{tabular}}
\end{table*}

\vspace{-3mm}
\paragraph{Medical history contribution} The impact of patient medical history on the quality of the Assessment model is depicted in Table \ref{table:assessment_performance}. To generate these results, models with different periods of medical history were trained (no history, half a year of history, and a full year of patient history). Across the board, the assessment model incorporating patient medical history information outperformed the model lacking such information, with a lift of $8.0\%$, $18.6\%$, $26.7\%$, and $80.9\%$ in PR-AUC for diagnoses, medications, labs, and imaging respectively.

\begin{table}
	\caption{Comparison of assessment model PR AUC, ROC AUC, and nDCG with different lengths of patient medical history}
	\label{table:assessment_performance}
	\centering
	\footnotesize
	\begin{tabular}{lllll}
\toprule
\makecell[l]{\textbf{Metric}} &
  \makecell[l]{\textbf{Target}}     & \textbf{No history} & \textbf{180 d}   & \textbf{365 d}   \\ \midrule
\multirow{4}{*}{PR AUC}  & Diagnoses & 0.400                       & 0.427                           & \textbf{0.432} \\
                      & Medications & 0.274 & 0.312 & \textbf{0.325} \\
                      & Labs        & 0.209 & 0.258 & \textbf{0.265} \\
                      & Imaging     & 0.105 & 0.174 & \textbf{0.190} \\ \midrule
\multirow{4}{*}{ROC AUC} & Diagnoses                      & 0.978                       & \textbf{0.979} & \textbf{0.979} \\
                      & Medications                      & 0.988 & 0.988 & \textbf{0.989} \\
                      & Labs                             & 0.933 & 0.943 & \textbf{0.944} \\
                      & Imaging                          & 0.854 & 0.869 & \textbf{0.875} \\
\multirow{4}{*}{nDGC} & Diagnoses                        & 0.582 & 0.595 & \textbf{0.597} \\  \midrule
                      & Medications                      & 0.229 & 0.233 & \textbf{0.235} \\
                      & Labs                             & 0.262 & 0.272 & \textbf{0.275} \\
                      & Imaging                          & 0.115 & 0.122 & \textbf{0.124}
                      \\ \bottomrule
\end{tabular}
\end{table}

\vspace{-2mm}
\section{Discussion}
\label{sec:discuss}

Here we have demonstrated SmartTriage, an automated system for engaging with patients virtually as a pre-visit questionnaire in order to provide documentation and decision support to their caregivers. Because SmartTriage was fully developed in-house, we were able to address two large short-comings of conventional symptoms checkers: (1) inability to utilize patient historical information and (2) lack of integration with an EMR. SmartTriage currently encodes up to a year of historical patient medical data, which includes demographic information, previous encounters, diagnoses, medications, procedures, laboratory results, and more. This information is used to avoid asking questions whose answers are already known (e.g. age, previous diagnoses), and uplifts the performance of all ML components utilized in SmartTriage. The ML components include the natural language understanding model for obtaining a chief complaint, the question sequencing model, and the assessment model for providing decision support.
 
Integrating with an EMR is crucial as it allows direct feedback from health care providers, and helps avoid unnecessary repetition for patients and gaps in continuity of care. SmartTriage’s EMR integration provides assurance that there is an easily understood record of the patient's interactions with the system while supporting clinicians in their task of generating documentation, orders, and diagnoses. Further discussion of the production integration is available in Appendix \ref{apd:prod}.  \figureref{fig:st_doc}, \ref{fig:st_assess} depict our main EMR integration, with a specially built activity that allows for manipulation of each reported patient response (\figureref{fig:st_doc}a) and clinical decision support (CDS) elements (\figureref{fig:st_assess}). \figureref{fig:st_doc}b shows what our documentation assistance looks like, with text written directly into the current progress note. The patient view of smart triage can be seen in \figureref{fig:st_patient_view}.

To create our pilot version of SmartTriage, we bootstrapped the system by using historical clinical encounters as a proxy for patient interactions -- we extracted patient reportable symptoms from clinical notes and employed the symptoms to stand in for possible answers. While we were able to obtain good decision support performance and clinically relevant question sequences, we recognize that the responses provided by patients may differ significantly from those in our training set in terms of interpretation (how the patient interprets the authored question) and overall question density (how many positive and negative examples are generated compared to the statistics found with NLP on clinical notes). We plan to continually retrain the system utilizing patient provided responses as more and more patients go through SmartTriage.

Although the ML components of SmartTriage were trained on healthcare data specifically from Southern California, the SmartTriage system itself would likely operate effectively in other health systems or regions. A significant effort was made to create comprehensive history of present illness rule-based (non-ML) questions to support documentation capture and quality of care. The chief complaints chosen for the initial version of SmartTriage are high frequency, primary care related, and typically have common, high evidence diagnostic and therapeutic outcomes. Considering the large volume of training data employed to generate the ML components of SmartTriage, tailoring the ML models to other systems or region specific populations would likely best be accomplished by fine-tuning on top of the existing trained models.
 
As the name implies, SmartTriage is meant to be a system for triaging patients. Our initial pilots and short-term roll-outs are an intermediate step to collecting the necessary ingredients for making high-quality triage decisions possible. By providing questionnaires to patients who have a forthcoming visit with a health care provider, we ensure that their data are appropriately reviewed, allowing us to learn how SmartTriage’s decision support performs in real encounter situations. Currently, all patients going through SmartTriage see a medical professional within 24 hours. Furthermore, as part of the EMR activity we developed, we ask caregivers to indicate what the most appropriate venue of care would have been for the visit, including self-care, a phone or video visit, an in-person primary care visit, urgent care, or the emergency department. The triage and decision support feedback will allow us to continually improve the SmartTriage system and ultimately position it as a tool for getting the right kind of care for the right kind of patient at the right time.
\vspace{-2mm}
\section*{Ethics}
This work was done as part of a quality improvement activity as defined in 45CFR §46.104(d)(4)(iii) -- secondary research for which consent is not required for the purposes of “health care operations” -- and as such ethical approval was not needed.
\vspace{-2mm}
\section*{Data availability}

Additional derived statistics about the data are available from the corresponding
author upon reasonable request. Raw patient data used to generate the models  is not publicly available.
\vspace{-2mm}
\section*{Acknowledgments}

This work would not be possible without the work and dedication of many individuals. We would like to thank Douglas Drugan, Scott Shebby, Taylor Kisor-Smith, Shlomit Kalcheim, Nadette Torres, Naqi Khan, Michael  Swiernik, Rommel Yabut, Michael Cairelli, Diane E. Oliver, Sergio Mendoza, Tanya Lozano, Ali Zaidi, Caleb Goodwin, Rajnish Sinha, Bayan Azima, Peter Li, Jason Ha, Naga Bramhananda Reddy, Vijaya ZakKula, Tulasi Kesanam, Danny Wudka, Jaime Wu, Kory Le, Calude Phan, Dhaval Patel, Navdeep Nagra, Michael Selter, Krishna Kumar Tadikamalla, Joshua Samaniego, Motez Musa, Krystal Patel, Matthew Yap, Onik Yeganian, Tian Jin, Felicia C Figuerres, Siao Jer, Betty Wan, Valery Vasquez, Jason Hermosa, and Jun Huang.

\bibliography{main.bib}

\counterwithin{figure}{section}
\counterwithin{table}{section}
\clearpage

\appendix

\section{Data sets }\label{apd:demo}

\begin{figure*}[!htb]
\centering
  
  \includegraphics[width=1\linewidth]{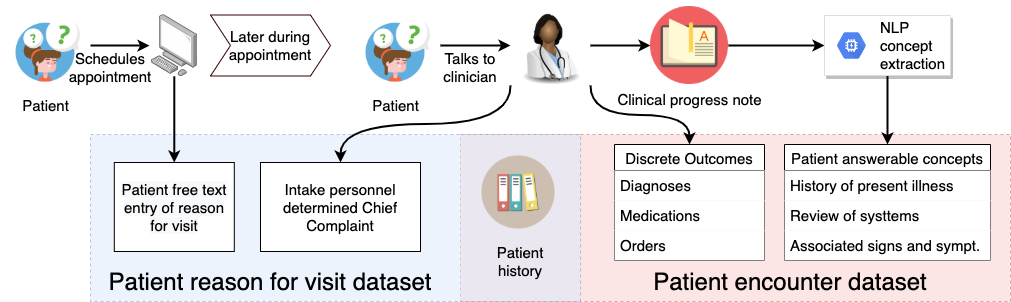}
  \caption{Diagram depicting how the patient reason-for-visit and patient encounter data sets are acquired. For the patient reason-for-visit data set, a free-text reason is entered when the appointment is scheduled. At the time of the encounter intake personnel choose a discrete chief complaint from a structured list. For the patient encounter data set, we combine discrete outcomes coded by the physician with NLP extracted concepts from clinical progress notes. Both data sets are further bundled with patient history elements including previous chief complaints, diagnoses, medications, laboratory results, and orders.}
  \label{fig:dataset_schematic}

\end{figure*}

\subsection{Reason-for-visit data set}
\label{sec:cc_demo}

We developed a Patient-Reason-for-Visit data set consisting of patient-generated reason-for-visit free-text entries, discrete chief complaints, demographic information, and patient historical information. The data set was constructed similarly to (anonymous citation) but with the addition of one year of historical information for each encounter. The data set was collected from \kaiser patients in Southern California who booked a Primary Care Physician appointment through our web portal, between February 2018 and June 2020. 

Patient medical history contained four information types: previous chief-complaints (954 distinct values), medications (1,276), diagnoses (1,497), and procedures (758). Medical history was limited to encounters that took place at most 365 days prior to the current visit. For each information source, the last 8 encounters containing relevant information were kept. The mean number of historic encounters with previous chief-complaint information is 5.8, where the mean number of chief-complaints per historic encounter was 1.5. For historic diagnoses the mean number of encounters with relevant information was 6.4, and the mean number of diagnoses per encounter was 2. For medication information there were 3.6 encounters on average, where each encounter contained 2 medications on average. For procedures the mean number of historic encounters was 4.2, with 1.6 relevant procedures on average.

Of the 1,296,210 encounters retrieved, there were 1,139,585 unique text entries that were assigned 2,050,851 chief complaints (of which 2,362 were unique). A subset of encounters had very generic patient text or non-specific chief complaints, and we used simple string matching to remove such examples. The resulting data set consisted of 907,170 encounters for 586,254 members. There were 708,782 unique free-text entries mapped to a total of 1,250,749 discrete chief complaints, 1,701 of which were unique. For our model development, we chose to classify a subset of 337 high frequency chief complaints. The median reason-for-visit text was 34 characters long, with a median word count of 6 (4 without stop words). For the reason-for-visit entries with additional free-text comments added by intake personnel, we cropped out comments to enforce the 50 character limit imposed by the web portal.

Among the patients in the final chief complaint data set, 335,476 (57\%) were female while 250,778 (43\%) were male. The average age was 49 years old. A full breakdown by age and sex can be found in Tables \ref{table:datasetAgeDistribution} and \ref{table:datasetSexDistribution}.

The data set was split into a training set of 816,357 encounters and 90,813 test encounters such that the two sets do not share the same patients or exact sentence matches. 

\begin{table*}[!h]
\caption{Age distribution of patients in the reason-for-visit data set. Percentage of total in parenthesis.}
\centering

\begin{tabular}{lll}
\toprule
\textbf{Age group} & \textbf{\# patients}    & \textbf{\# encounters}     \\
\midrule
0 - 1                 & 15,050 (2.5\%)  & 28,037        (3.1\%)  \\
2 - 15                & 49,356 (8.3\%)  & 81,108       (8.9\% ) \\
16 - 20               & 14,178 (2.4\%)  & 19,306        (2.1\% ) \\
21 - 29               & 98,496  (16.5\%) & 144,646       (15.9\%) \\
30 - 39               & 121,733     (20.4\%) & 184,205       (20.3\%) \\
40 - 49               & 89,399      (15.0\%) & 136,482       (15.0\%) \\
50 - 59               & 81,047      (13.6\%) & 120,824       (13.3\%) \\
60 - 69               & 75,059      (12.6\%) & 113,470       (12.5\%) \\
70 - 79               & 37,867      (6.4\%)  & 58,310        (6.4\%)  \\
80 - 89               & 11,347      (1.9\%)  & 17,580        (1.9\%)  \\
90+                   & 2,160       (0.4\%)  & 3,202         (0.4\%) \\
\midrule
Total                 & 595,692             & 907,170              \\
\bottomrule 
\end{tabular}
\label{table:datasetAgeDistribution}
\end{table*}

\begin{table*}[!h]
\caption{Biological sex distribution of patients in the reason-for-visit data set. Percentage of total in parenthesis.}
\centering

\begin{tabular}{@{}lll@{}}
\toprule
\textbf{Biological sex} & \textbf{\# patients} & \textbf{\# encounters} \\ \midrule
Female                  & 335,476 (57.22\%)    & 534,940 (58.97\%)      \\
Male                    & 250,778 (42.78\%)    & 372,230 (41.03\%)      \\
Total                   & 586,254              & 907,170                \\ \bottomrule
\end{tabular}
\label{table:datasetSexDistribution}
\end{table*}

\subsection{Encounter data set}
\label{sec:enc_demo}

To create the content for our platform we employed a strategy that leveraged our organization’s large patient membership. The primary input knowledge source was a set of 25,047,535 historical primary care encounters with chief complaints, clinical notes and patient demographics. Our output data source consisted of the resulting ICD10 diagnoses, laboratory orders, imaging studies, and medications that were associated with the encounters. 

The clinical notes we extracted are largely unstructured text documents that contain roughly the same general categories of information - why the patient was being seen, a brief medical history, a review of systems and symptomatology, a focused physical exam, an assessment, and a plan. To access these critical elements of the patient's presentation we processed all of the notes to extract clinical concepts using an in-house developed NLP engine. The engine incorporates, among other sources, the UMLS (Unified Medical Language System) terminology. 

Each note yielded between 10-200 (or more) medical history items, symptoms, physical exam findings, and some elements of patient social and family history. The NLP engine assigns every item a probability that it is present (positive) or absent (negative) and identifies the section in the note that the item came from. For the purposes of SmartTriage we limited ourselves to considering items from the Chief Complaint (CC), History of Present Illness (HPI), Review of Systems (ROS), and Physical Exam (PE) sections, and only considered positive or negative items where the classification probability was $>= 90\%$. 

The resulting 25 million encounter records annotated 4,110,052 patients with unique 2,900 chief complaints and 24,472 ICD-10-CM diagnose codes. Tables \ref{table:encountersStatsByAge} and \ref{table:encountersStatsBySex} show the full breakdown of HPI, medical history records, and clinical concepts by gender and age group.

\begin{table*}[]
	\caption{Age segmented statistics for the encounters data set}
	\label{table:encountersStatsByAge}
	\begin{tabular}{@{}llllll@{}}
		\toprule
		\textbf{Age Group} & \textbf{Historical}     & \textbf{UMLS Concepts}   & \textbf{Severity} \& \textbf{Duration} & \textbf{HPI Concepts} & \textbf{Total}  \\ \midrule
		2 - 15     & 18368(15.71\%) & 75458(64.53\%)  & 21715(18.57\%)       & 1398(1.20\%) & 116939 \\
		16 - 20    & 33699(18.25\%) & 113872(61.67\%) & 35489(19.22\%)       & 1589(0.86\%) & 184649 \\
		21 - 29    & 47743(20.48\%) & 138151(59.25\%) & 45572(19.55\%)       & 1684(0.72\%) & 233150 \\
		30 - 39    & 56880(21.25\%) & 155805(58.22\%) & 53195(19.88\%)       & 1755(0.66\%) & 267635 \\
		40 - 49    & 65236(22.22\%) & 168297(57.33\%) & 58240(19.84\%)       & 1794(0.61\%) & 293567 \\
		50 - 59    & 69826(22.63\%) & 176086(57.07\%) & 60807(19.71\%)       & 1828(0.59\%) & 308547 \\
		60 - 69    & 72559(23.51\%) & 174957(56.69\%) & 59299(19.21\%)       & 1823(0.59\%) & 308638 \\
		70 - 79    & 65289(23.19\%) & 161833(57.48\%) & 52653(18.70\%)       & 1755(0.62\%) & 281530 \\
		80 - 89    & 49265(21.82\%) & 134266(59.45\%) & 40657(18.00\%)       & 1641(0.73\%) & 225829 \\
		90 +  & 31415(20.09\%) & 96605(61.77\%)  & 26882(17.19\%)       & 1492(0.95\%) & 156394 \\ \bottomrule
	\end{tabular}

\end{table*}

\begin{table*}[!htb]
	\caption{Biological sex segmented statistics for the Encounters data set}
	\label{table:encountersStatsBySex}
		\begin{tabular}{@{}llllll@{}}
		\toprule
		\textbf{Biological sex} & \textbf{Historical}      & \textbf{UMLS Concepts}   & \textbf{Severity \& Duration} & \textbf{HPI Concepts} & \textbf{Total}   \\ \midrule
		Female         & 286842(22.63\%) & 729148(57.53\%) & 242815(19.16\%)      & 8517(0.67\%) & 1267322 \\
		Male           & 223438(20.14\%) & 666182(60.04\%) & 211694(19.08\%)      & 8242(0.74\%) & 1109556 \\ \bottomrule
	\end{tabular}

\end{table*}

\newpage

\section{Production integration}
\label{apd:prod}

\begin{figure}[]
	\centering
	\includegraphics[width=0.8\linewidth]{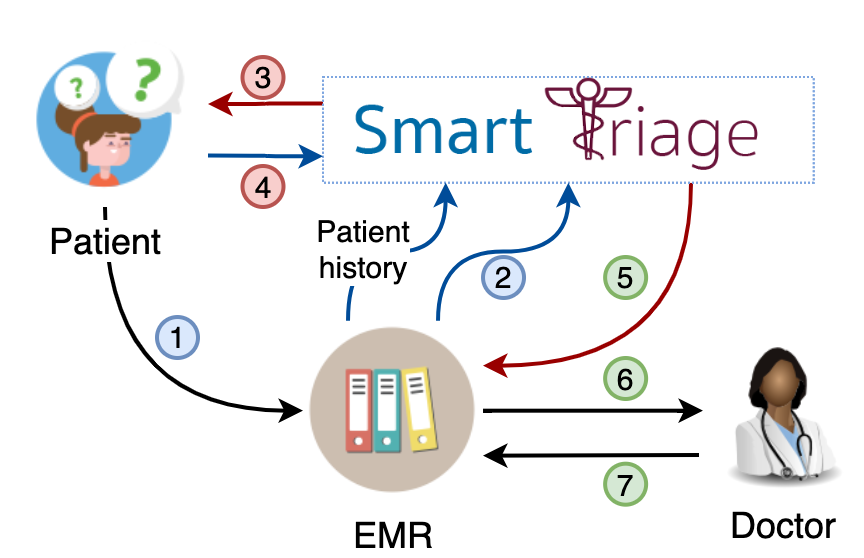}
	\caption{Interaction overview diagram for SmartTriage. 
	1-2. Appointment is created, SmartTriage extracts all new appointments once per hour.
	3-4. SmartTriage interacts with patient to obtain encounter specific information. 
	5. SmartTriage inserts patient assessment back into the EMR. 
	6. Provider views patient information and assessment
	7. Further actions by the provider and feedback on SmartTriage are recorded.}
\label{fig:interaction-diagram}
\end{figure}

SmartTriage is currently being piloted across the \kaiser Southern California Region in virtual visits (phone, video) as a pre-visit symptom checker. If the pilot is successful, SmartTriage will likely be rolled out widely throughout our health care organization, impacting millions of patients. To enable production deployment, we had to build multiple interfaces with existing enterprise systems. A high-level interaction diagram is presented in \figureref{fig:interaction-diagram}.

\paragraph*{ETL process for patient history} Rather than directly query the EMR, we maintain a local copy of the discrete patient history information for members in our geographic region. We refresh our copy ($\sim 500$ million rows) as a nightly extraction process.

\paragraph*{Encounter start notification} We extract all new appointments every hour and send email notification 24 hours prior to the appointment. If the patient does not engage within 3 hours of the appointment, we send a text message to their mobile device.

\begin{figure*}[!htb]
	\centering
	\includegraphics[width=0.5\linewidth]{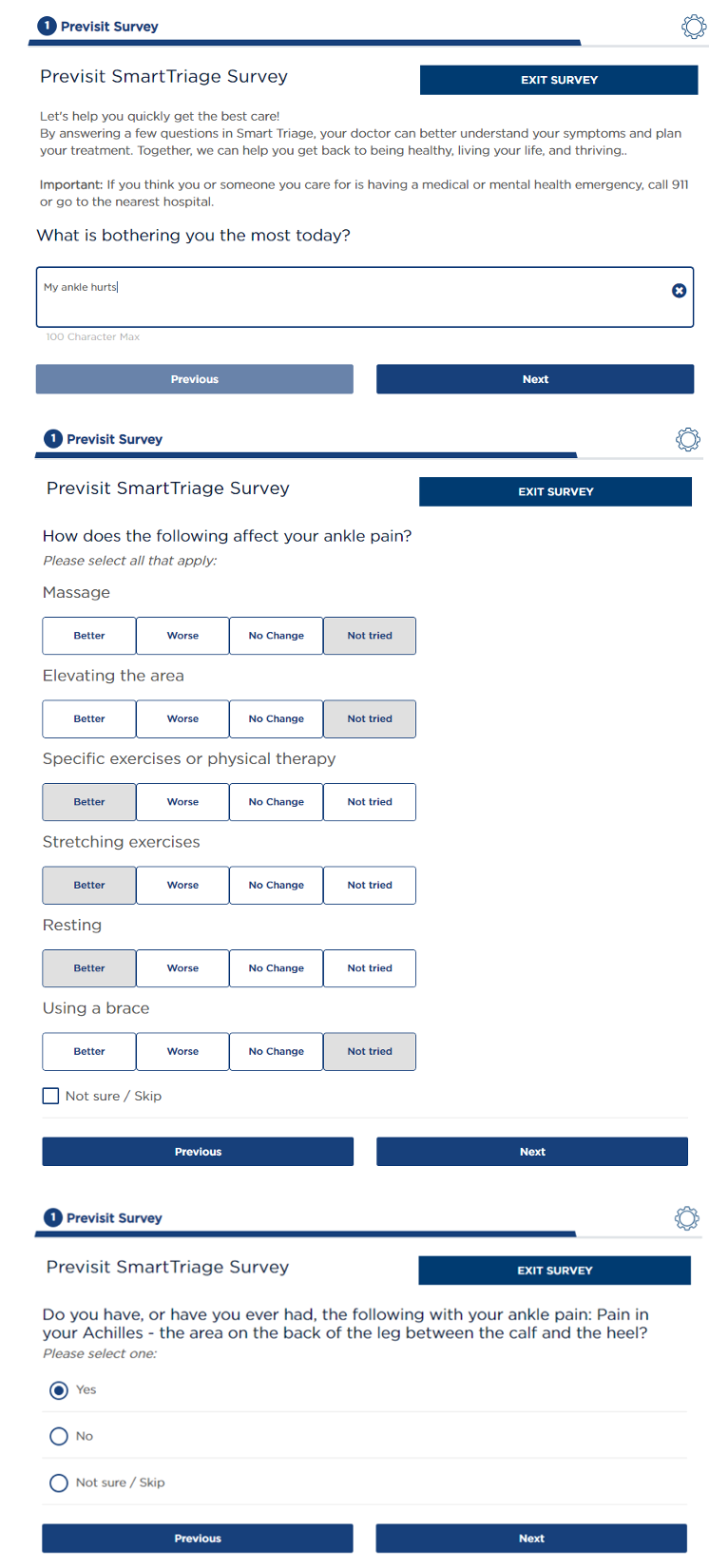}
	\caption{Screen shots from the patient view of smart triage.}
	\label{fig:st_patient_view}
\end{figure*}

\begin{figure*}[!htb]
  \centering
  \includegraphics[width=1\linewidth]{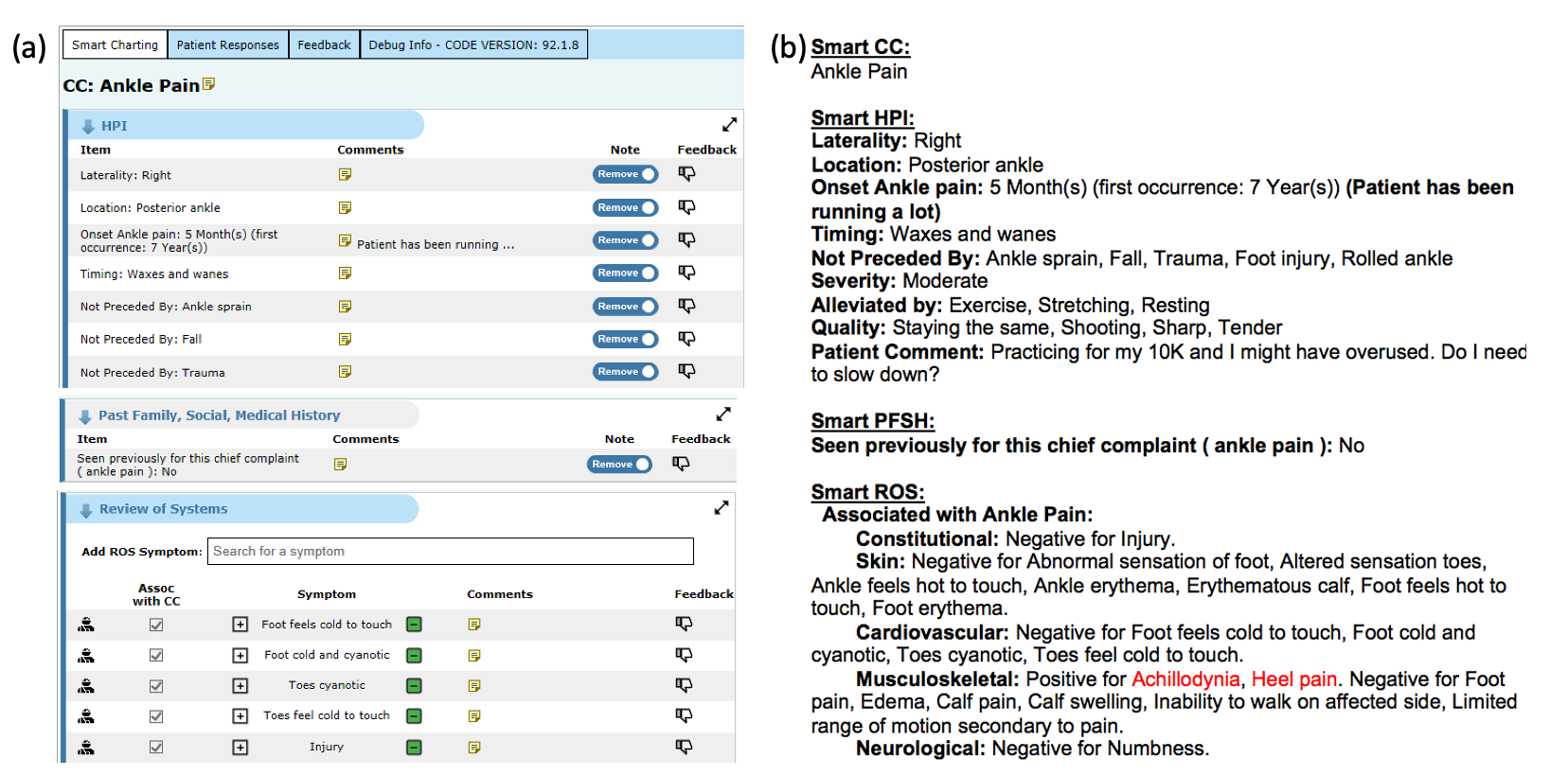}
  \caption{Provider view of SmartTriage. (a) The EMR activity is what the clinician sees when they open the patient’s encounter. Each discrete piece of captured data is actionable. (b) The EMR note is dynamically generated by the data in the EMR activity. Positive symptoms are in red.}
  \label{fig:st_doc}
\end{figure*}

\begin{figure*}[!htb]
  \centering
  \includegraphics[width=0.4\linewidth]{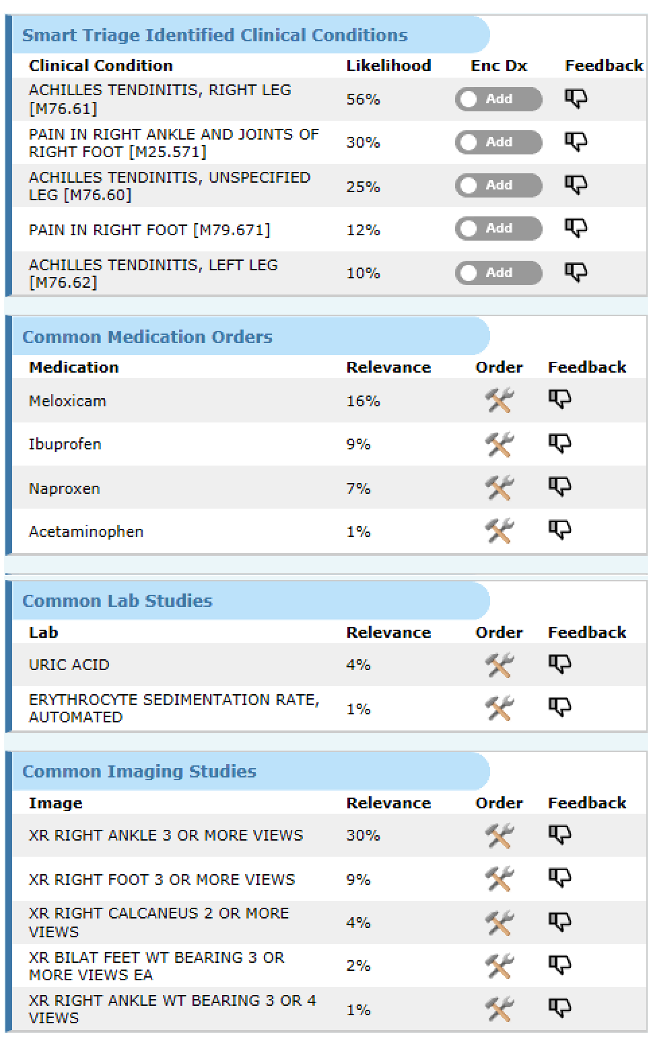}
  \caption{Provider view of SmartTriage’s Assessment activity. Recommendations for encounter diagnoses, medications, labs, and imaging studies are provided. Diagnoses can be directly added into the current encounter. }
  \label{fig:st_assess}
\end{figure*}

\paragraph*{Webapp API} We implemented a custom REST API to communicate with the enterprise web UI service. When the patient clicks on the link in the email or text message they are redirected to the the enterprise patient authentication service, and then presented with the patient view of SmartTriage (\figureref{fig:st_patient_view}).  
\paragraph{Inputting data back into the EMR} We leverage existing SOAP APIs to send the patient data back into the EMR.
\paragraph{Provider facing interface} We implemented a custom UI in the EMR for the clinicians to interact with the SmartTriage data, integrate into patient chart, and provide feedback to the SmartTriage system (see \figureref{fig:st_doc}, \ref{fig:st_assess}).
\paragraph{Daily reporting} A daily reporting job is used to extract clinician feedback and collect data for re-training of the machine learning models.

\clearpage
\onecolumn
\section{Content management system}

\begin{figure*}[htbp]
	\centering
	\includegraphics[width=0.9\linewidth]{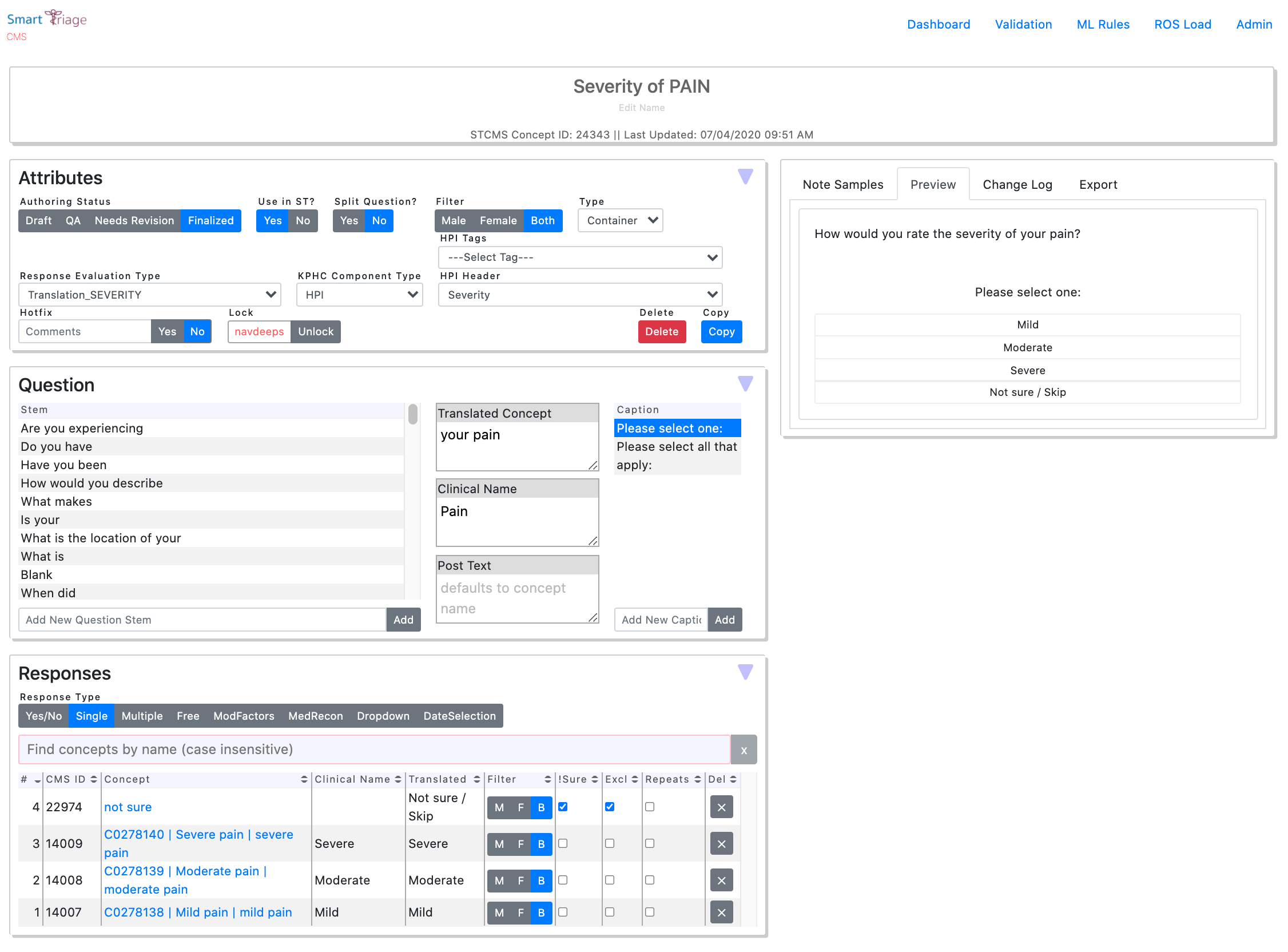}
	\caption{Screen shot of the content management system. Concepts can be grouped according to synonymy and authored for the patient (``Translated concept") and EMR (``Clinical name"). UI components are built-in to help author meta-data for both the question as it appears to the patient (question type, stem, etc.) and as it appears to the EMR (component type, system name or header name when applicable, etc.). The assertion status determines how the answer(s) to the question is/are formulated when sent to the ML models.}
	\label{fig:cmsview}
\end{figure*}

\newpage
\section{Machine Learning Component Architecture}

\begin{figure*}[!htb]
	\centering
	\includegraphics[width=0.7\linewidth]{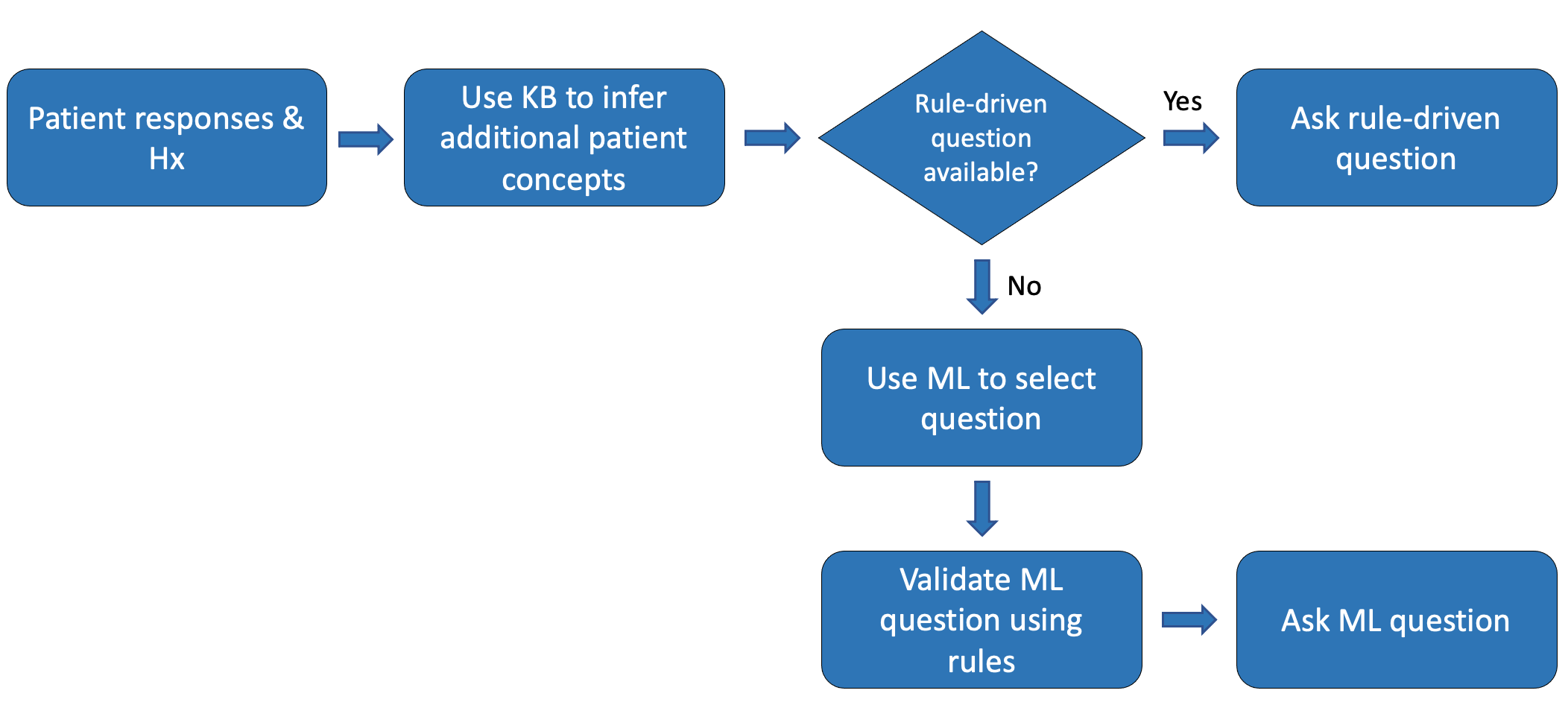}
	\caption{Schematic of a single round in the question selection process.}
	\label{fig:question_selection}
\end{figure*}

\begin{figure*}[htbp]
     \floatconts{fig:neural_net_components}
     {\caption{Neural network components of SmartTriage \textbf{a.} Chief Complaint Model Architecture \textbf{b.} Assessment Model Architecture \textbf{c.} Medical history embedding }} 
     {
     \includegraphics[width=0.9\textwidth]{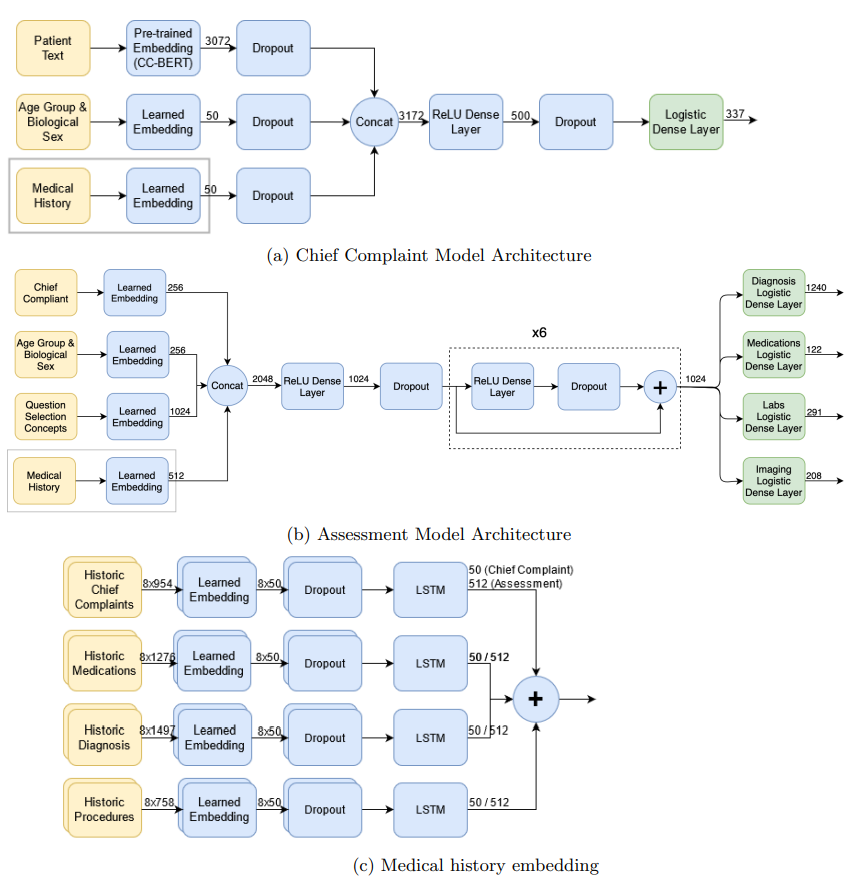}
     \label{fig:assessment_model_arch}
   
}
      
\end{figure*}

\begin{table*}[h!]
	\caption{Chief complaint model performance on validation set comparing hyperparameters}
	\centering
	\begin{tabular}{ll|lll}
		\midrule
		Hidden layer size & Dropout rate & \textbf{PR AUC}        & \textbf{ROC AUC}       & \textbf{nDCG} \\
		\toprule
		100 & 0.0 & 0.4356±0.0034 & 0.9242±0.0002 & 0.7086±0.0015 \\
		100 & 0.3 & 0.4311±0.0026 & 0.9263±0.0003 & 0.7045±0.0016 \\
		100 & 0.5 & 0.4084±0.0050 & 0.9245±0.0001 & 0.6897±0.0036 \\
		500 & 0.0 & 0.4423±0.0005 & 0.9226±0.0015 & 0.7105±0.0007 \\
		500 & 0.3 & 0.4535±0.0005 & 0.9307±0.0006 & 0.7169±0.0003 \\
		500 & 0.5 & 0.4508±0.0009 & 0.9326±0.0003 & 0.7151±0.0003 \\
		1000 & 0.0 & 0.4407±0.0001 & 0.9230±0.0013 & 0.7103±0.0001 \\
		1000 & 0.3 & 0.4505±0.0007 & 0.9325±0.0002 & 0.7152±0.0001 \\
		1000 & 0.5 & \textbf{0.4549±0.0001} & 0.9344±0.0001 & \textbf{0.7172±0.0001} \\
		2000 & 0.0 & 0.4406±0.0003 & 0.9246±0.0008 & 0.7104±0.0001 \\
		2000 & 0.3 & 0.4447±0.0006 & 0.9330±0.0022 & 0.7123±0.0002 \\
		2000 & 0.5 & 0.4535±0.0005 & \textbf{0.9351±0.0004} & 0.7169±0.0003 \\		
		\bottomrule   
		\label{table:CCHyperPerformance}
	\end{tabular}
\end{table*}

\end{document}